\pdfoutput=1

\documentclass[11pt]{article}

\usepackage[]{acl}

\usepackage{times}
\usepackage{latexsym}
\usepackage{color,soul}
\usepackage[T1]{fontenc}

\usepackage[utf8]{inputenc}

\usepackage{microtype}
\usepackage{amsmath}
\usepackage{graphicx}
\title{Flow-Adapter Architecture for Unsupervised Machine Translation}

\author{Yihong Liu$^{1}$, Haris Jabbar$^2$ \and Hinrich Schütze$^2$\\
$^1$Department of Informatics,
Technical University of Munich \\
   $^2$Center for Information and Language Processing, LMU Munich\\
  \texttt{yihong.liu@tum.de} \\
   \texttt{jabbar@cis.lmu.de} \\
   }

\newcounter{notecounter}
\newcommand{\enotesoff}{\long\gdef\enote##1##2{}}

\enotesoff                      

\def\myparagraph#1{\textbf{#1}}

\begin{document}
\maketitle

\begin{abstract}
In this work, we propose a \emph{flow-adapter architecture} for unsupervised NMT. It leverages normalizing flows to explicitly model the distributions of sentence-level latent representations, which are subsequently used in conjunction with the attention mechanism for the translation task. The primary novelties of our model are: (a) capturing language-specific sentence representations separately for each language using normalizing flows and (b) using a simple transformation of these latent representations for translating from one language to another. This architecture allows for unsupervised training of each language independently. While there is prior work on latent variables for supervised MT, to the best of our knowledge, this is the first work that uses latent variables and normalizing flows for unsupervised MT. We obtain competitive results on several unsupervised MT benchmarks.
\end{abstract}

\enote{hs}{don't forget to reove hypersetup draft}
\enote{yl}{sure, we will do that for the final version}


\section{Introduction}

Recent advances in deep learning have boosted the
development of neural machine translation (NMT). Typical NMT
models leverage an encoder-decoder
framework \citep{cho2014learningphrase,sutskever2014sequence2sequence}. However,
NMT models have been shown to be data-hungry, as the number
of parallel sentences significantly influences the
performance \citep{zoph2016transfer}. Unfortunately,
large-scale bilingual corpora are limited to a relatively
small subset of languages \citep{al2002translation}. In contrast to bilingual corpora, monolingual corpora are much easier to obtain. 

\begin{figure}
  \centering
  \includegraphics[width=0.4\textwidth]{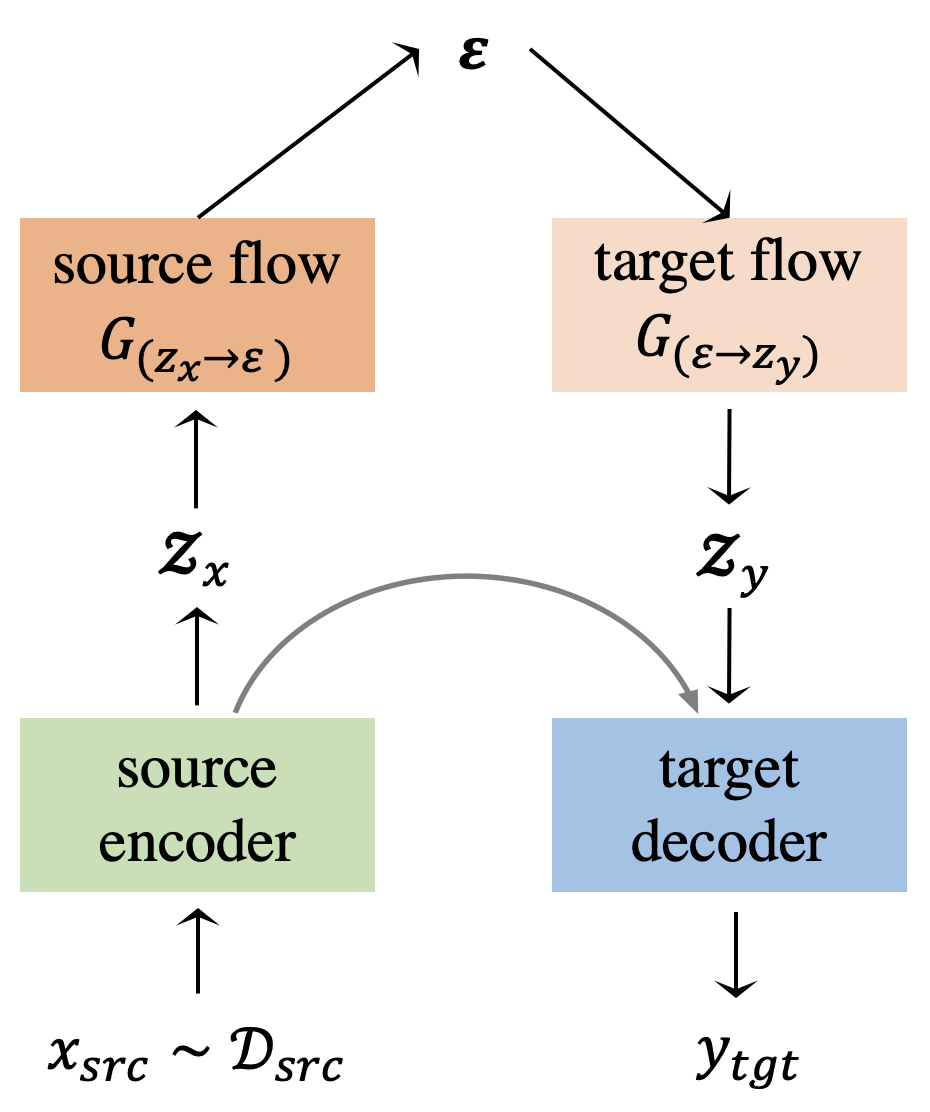}
  \caption{Inference pipeline of proposed flow-adapter based model for source-to-target translation. The decoder also uses the attentional input (shown as the gray arrow between the encoder and the decoder).} \label{fig:flow-adapter_structure}
\end{figure}

\enote{hj}{Added this graphic instead of Fig 1 to give an overview of proposed architecture. What do you guys think? yl: Refer my email. Need improved quality }
\enote{yl}{I updated the figure.}

Unsupervised NMT, compared with supervised NMT, aims to
train a model without parallel data. Some early
works \citep{irvine2016end,sennrich2016improvingn,cheng2016semisupervised}
used monolingual corpora to boost performance when parallel
data is not abundant. \citet{LampleCDR18umtmco}
and \citet{ArtetxeLAC18unmt} explored the possibility of
training a model relying only on monolingual corpora.  They
both leveraged a shared-encoder architecture in order to
generate universal representations, trained with techniques
such as
initial word-by-word
translation
through bilingual dictionaries
\citep{LampleCRDJ18wordtrans,artetxe2017learningbi},
denoising auto-encoding (DAE) \citep{vincent2008extracting}
and iterative back-translation
(BT) \citep{HoangKHC18ibt}. However, \citet{yang2018unsupervised}
argued that it is a bottleneck in such shared-encoder models to use a
shared encoder that maps pairs of sentences of different
languages to the same shared latent space. They proposed to use two
independent encoders sharing part of their weights and
achieved better results.
But all of those aforementioned approaches trained the
translation models almost from scratch (with only some prior
knowledge in the pre-trained embeddings) and therefore it is
hard to further advance their performance.

More recently, with the advance in pre-trained
models \citep{peters2018deepcontext,devlin2019bert},
researchers have begun to explore the possibility of using
pre-trained models for unsupervised
NMT. \citet{Conneau2019xlm} extended the pre-training from a
single language to multiple languages, referred to as
cross-lingual pre-training. By using  pre-trained
cross-language models (XLMs) to initialize  encoder and
decoder, they achieved good unsupervised MT
performance on multiple language
pairs. In related work, \citet{SongTQLL19mass} 
proposed masked sequence to sequence pre-training
(MASS), which  directly pre-trains a whole
encoder-decoder model. \citet{ustun2021multilingual}
proposed a language-specific denoising-adapter architecture
to increase the multilingual modeling capacity of the
pre-trained model mBART \citep{liu2020mbart} and used these
adapters for multilingual unsupervised NMT. Although these
adapters are trained with monolingual data only, the
fine-tuning step relies on parallel data.

Current NMT frameworks rely heavily on the attention
mechanism \citep{bahdanau2015neuralalign,vaswani2017attention}
to capture alignments. However,  attention-based
context vectors can fail to extract sufficiently accurate
sentence-level semantics and thus result in incorrect
translations
or translation
ambiguity \citep{tu2016modeling, zhang2016variationalnmt}.
To tackle this issue,
several variational frameworks for modeling the translation
process have been proposed
\citep{zhang2016variationalnmt, eikema2019auto, Hendra2020variational}. These
approaches incorporate sentence-level latent representations
into NMT.
A latent representation, in the context of this
paper, is a fixed-size continuous vector from an unknown distribution that captures the
semantics of a source sentence. The
target sentence is then generated from
this latent representation using a simple transformation along with the attention mechanism commonly found in transformer architectures. In this way,
when the attention mechanism learns incorrect alignments, the
latent representation plays a complementary role in
guiding the translation.

Prior work in this vein has only been conducted in
supervised NMT.
In this paper,
 we propose a flow-adapter architecture for unsupervised
 NMT. Similar to variational methods, we  model the
 distribution of sentence-level representations. However,
 unlike  variational methods, which model the distribution
 in an implicit way, we use a pair of normalizing flows to
 explicitly model the distributions of source and target
 languages. Secondly, different from some previous
 unsupervised NMT models that assume that the
 representations of source and target sentences share a
 common semantic space, we assume the representations are
 different because of language-specific
 characteristics. Hence they are modeled separately for each language. Subsequently a simple transformation converts source representations
 into target representations.
This makes it possible to
better capture sentence semantics in a
 language-specific manner.
Lastly, instead of minimizing KL loss,
the flows are directly trained by maximum likelihood estimation (MLE) of sentence-level latent representations. This gives the latent representations more flexibility. 

Our main contributions:\\
(1) We propose a novel flow-adapter architecture. It
uses normalizing flows to explicitly model the distributions
of sentence-level representations and performs a latent representation
transformation from source to target.
To the best of our knowledge, this is the first work that uses latent variables and normalizing flows for unsupervised NMT.\\
(2) Experiments show the validity and effectiveness
of our flow-adapter architecture. It performs very well in
unsupervised NMT on several language pairs on the Multi30K
dataset. When additionally using pre-trained models, we
achieve  results competitive with the state of the art on WMT
datasets, especially for \textit{en-fr} (WMT'14)
and \textit{en-ro} (WMT'16).

\section{Background}

\subsection{Normalizing Flows}

Normalizing flows (NFs) are a special type of deep
generative model. Different from generative adversarial
networks (GAN) \citep{Goodfellow2014GAN} and variational
auto-encoding (VAE) \citep{Kingma2014Auto}, NFs allow for
not only sampling but also exact density estimation. Due to
such desirable properties, in recent years, they have been
successfully applied to fields such as
image \citep{Ho2019flowplus,Kingma2018Glow},
audio \citep{esling2019universal,Parallel2018van} and video
generation \citep{kumar2019videoflow}. In addition to
significant achievements in modeling continuous data, NFs
have also been used for modeling discrete data, 
either by directly modeling the data in discrete
space \citep{tran2019discrete, hesselink2020latent} or
by transforming the discrete data into continuous
space \citep{zachary2018latentnormalizing,
tang2021continuous}.

NFs transform between two distributions based on the
following change-of-variables formula (we follow the introduction of \citep{dinh2015nice,dinh2017realnvp}):
\begin{equation}
\log p_x(\boldsymbol{x}) = \log p_z(\boldsymbol{z}) + \log \left| \det \frac{\partial f_{\theta}(\boldsymbol{z})}{\partial \boldsymbol{z}} \right|^{-1}
\label{densitytransform1}
\end{equation}
where $\boldsymbol{z} \sim p_z(\boldsymbol{z})$ and
$\boldsymbol{x} \sim p_x(\boldsymbol{x})$ denote two vectors
from a simple latent distribution $p_z(\boldsymbol{z})$ and
a complex distribution of the observed data $p_x(\boldsymbol{x})$, $f_{\theta}$ is an
invertible and differentiable function (neural network with parameters $\theta$), $f_{\theta}(\boldsymbol{z}) = \boldsymbol{x}$ and 
$\det \frac{\partial
f_{\theta}(\boldsymbol{z})}{\partial \boldsymbol{z}}$ denotes the
determinant of the Jacobian matrix of $f_{\theta}$. The idea of NFs
is to learn an $f_{\theta}$ such that $f_{\theta}$ and
$f_{\theta}^{-1}$  transform between
the latent space
$p_z(\boldsymbol{z})$ and the observed space $p_x(\boldsymbol{x})$.

Constructing a single arbitrarily complex invertible and
differentiable function is usually cumbersome. Therefore, a
generally adopted approach is to stack multiple
transformations $f_i$ together, i.e., $\boldsymbol{x} =
f_{\theta}(\boldsymbol{z}) = f_K \circ \cdots \circ
f_1(\boldsymbol{z})$. Similarly, for the reverse direction we
have $\boldsymbol{z} = f_{\theta}^{-1}(\boldsymbol{x}) =
f_1^{-1} \circ \cdots \circ f_K^{-1}(\boldsymbol{x})$, whose
Jacobian matrix is efficient to compute. Here $K$ denotes
the number of sequential flows (e.g., $K=3$ in
Table \ref{tab:multi_trans_perf}).



Normalizing flows are usually optimized by MLE of the
parameters $\theta$, i.e., $\log p(\mathcal{D}|\theta)
= \sum_{n=1}^{N} \log p_x (\boldsymbol{x}^{(n)} | \theta)$,
where $N$ is the data size. By applying a variant of the change-of-variable formula in Equation (\ref{densitytransform1}), i.e., $\log p_x(\boldsymbol{x}) = \log p_z( f_\theta^{-1}(\boldsymbol{x})) + \log \left| \det \frac{\partial f_\theta^{-1}(\boldsymbol{x})}{\partial \boldsymbol{x}} \right|$, the MLE objective can be reformulated as follows:
\begin{equation}
\begin{aligned}
\log p(\mathcal{D}|\theta) & = \sum_{n=1}^{N} \log p_z(f_\theta^{-1}(\boldsymbol{x}^{(n)})|\theta) \\
& \,\,\,\,\,\,\,\, + \log \left| \det \frac{\partial f_\theta^{-1}(\boldsymbol{x}^{(n)})}{\partial \boldsymbol{x}^{(n)}} | \theta \right|
\end{aligned}
\label{mle}
\end{equation}

\subsection{Latent-variable (variational) NMT}

Compared with standard encoder-decoder based NMT models, latent-variable (variational)  approaches \citep{zhang2016variationalnmt, eikema2019auto, flowseq2019, calixto2019latentmmt, Hendra2020variational, latent2020shu} additionally leverage latent random variables.

Let $\boldsymbol{x}$ be a sentence from the source language and $\boldsymbol{y}$ be its translation in the target language. Then, the variational NMT framework introduces a continuous random latent variable $\boldsymbol{z}$ for the translation modeling, i.e., $p(\boldsymbol{y}| \boldsymbol{z}, \boldsymbol{x})$. With the introduction of $\boldsymbol{z}$, the conditional probability $p(\boldsymbol{y}| \boldsymbol{x})$ can then be reformulated as follows:
\begin{equation}
p(\boldsymbol{y} | \boldsymbol{x}) = \int_{\boldsymbol{z}} p(\boldsymbol{y}|\boldsymbol{z}, \boldsymbol{x}) p(\boldsymbol{z}|\boldsymbol{x}) d{\boldsymbol{z}}
\label{conditional_prob}
\end{equation}

In this way, $\boldsymbol{z}$ serves as a global semantic
signal that is helpful to counteract incorrect alignments
the model has learned and uses through attention.  However,
the integration of $\boldsymbol{z}$ poses challenges for
inference. To address this problem, variational NMT adopts
techniques from
VAE \citep{Kingma2014Auto,rezende2014stochasticback},
namely, neural approximation and the reparameterization
trick.

Neural approximation leverages a neural network to approximate the posterior distribution $p(\boldsymbol{z} | \boldsymbol{x}, \boldsymbol{y})$ with $q_{\phi}(\boldsymbol{z} | \boldsymbol{x}, \boldsymbol{y})$, where $\phi$ denotes the parameters of the neural network. In most works, $q_{\phi}(\boldsymbol{z} | \boldsymbol{x}, \boldsymbol{y})$ is designed as a diagonal Gaussian $\mathcal{N}(\boldsymbol{\mu}, \text{diag}(\boldsymbol{\sigma}^2))$, where the mean $\boldsymbol{\mu}$ and the variance $\boldsymbol{\sigma}^2$ are parameterized with neural networks.

Reparameterization means that the latent random
variable $\boldsymbol{z}$ is parameterized as a function of the mean
$\boldsymbol{\mu}$ and the variance
$\boldsymbol{\sigma}^2$. In this way, the gradient with
respect to the parameters $\boldsymbol{\mu}$ and
$\boldsymbol{\sigma}^2$ can be computed. The
reparameterization of $\boldsymbol{z}$ is often carried out in a
location-scale manner: $\boldsymbol{z} = \boldsymbol{\mu}
+ \boldsymbol{\sigma} \odot \boldsymbol{\epsilon}$ where $\boldsymbol{\epsilon} \sim \mathcal{N}(0,1)$

With these two techniques, the learning objective of variational NMT is the evidence lower-bound or ELBO of the conditional probability $p(\boldsymbol{y}| \boldsymbol{x})$:
\begin{equation}
\begin{aligned}
\mathcal{L}(\theta, \phi; \boldsymbol{x}, \boldsymbol{y}) &= - \text{KL}(q_{\phi}(\boldsymbol{z} | \boldsymbol{x}, \boldsymbol{y})||p_{\theta}(\boldsymbol{z} | \boldsymbol{x})) \\ 
& \,\,\,\,\,+ E_{q_{\phi}(\boldsymbol{z} | \boldsymbol{x}, \boldsymbol{y})}[\log p_{\theta}(\boldsymbol{y} | \boldsymbol{z}, \boldsymbol{x})]
\label{vnmt_objective}
\end{aligned}
\end{equation}
where $p_{\theta}(\boldsymbol{z} | \boldsymbol{x})$ is the
prior distribution modeled  by a neural network and
$p_{\theta}(\boldsymbol{y}
| \boldsymbol{z}, \boldsymbol{x})$ is modeled by the decoder
given the input source sentence $\boldsymbol{x}$ and the
latent variable $\boldsymbol{z}$. The KL term minimizes the
discrepancy between the prior $p_{\theta}(\boldsymbol{z}
| \boldsymbol{x})$ and the posterior
$q_{\theta}(\boldsymbol{z}
| \boldsymbol{x}, \boldsymbol{y})$. In the inference step,
$\boldsymbol{z}$ can therefore be sampled from the prior,
which only requires $\boldsymbol{x}$ instead of the
posterior that requires both $\boldsymbol{x}$ and
$\boldsymbol{y}$. Although this variational framework
leverages latent variables, which are helpful for translation, it still has some flaws: \textbf{1)} training a variational NMT framework requires parallel corpora to construct the posterior $q_{\phi}(\boldsymbol{z} | \boldsymbol{x}, \boldsymbol{y}$) and such parallel corpora are not available for unsupervised MT; \textbf{2)} the distribution family of the latent variables, e.g., $p_{\theta}(\boldsymbol{z} | \boldsymbol{x})$, is pre-defined, e.g., a Gaussian, which might restricts the advantage of using a complex posterior; \textbf{3)} as variational NMT leverages $\boldsymbol{z}$ sampled from $p_{\theta}(\boldsymbol{z} | \boldsymbol{x})$ for inference, an underlying assumption is that $\boldsymbol{z}$ should be the same whether only $\boldsymbol{x}$ is considered or both $\boldsymbol{x}$ and $\boldsymbol{y}$ are considered. In other words, this framework assumes $\boldsymbol{z}$ is language-agnostic, which might not be true since language-specific characteristics can influence the generation of $\boldsymbol{z}$.


\section{Flow-Adapter Based Framework}

In this work, we want to reap the benefits of introducing
latent variables into unsupervised MT while at the
same time avoiding
the flaws of variational NMT we just discussed. Therefore, we
propose a flow-adapter based framework that uses two NFs
to explicitly model the distribution of the sentence-level
latent representations of the source and target
sentences. In this way, we can take account of multilinguality in
unsupervised MT and make use of language-specific
sentence-level representations. During the translation
process, a latent code transformation is performed to
transform the source-language representation into
the target-language representation so that the decoder can
leverage them to generate a better target-language
sentence. We will first introduce the sentence-level
representation as well as the latent code transformation in
Section \ref{latentcode}, followed by the description of the
flow-adapter based framework for unsupervised MT in
Section \ref{flowUMNT}.

\subsection{Modeling Representation by NFs \& Latent Code Transformation}\label{latentcode}

As previously mentioned, variational methods such
as \citep{zhang2016variationalnmt,Hendra2020variational}
assume that the semantics of the source sentence
$\boldsymbol{x}$ and target sentence $\boldsymbol{y}$ are
the same and thus the generated latent variable
$\boldsymbol{z}$ is the same regardless of whether we  only
consider $\boldsymbol{x}$ or consider both $\boldsymbol{x}$ and
$\boldsymbol{y}$. Unsupervised NMT methods such
as \citep{LampleCDR18umtmco,Conneau2019xlm}  similarly
assume that a shared encoder maps source and target
sentences into a shared latent space.

In this work, however, we diverge from this assumption and
follow \citet{yang2018unsupervised} in adopting the
desideratum that
the unique and internal characteristics of each
language be respected. One could think that the semantics of a pair of sentences
should theoretically be the same; but in reality,
 because of
language-specific characteristics, the latent representations
$\boldsymbol{z}$ obtained by an encoder can be different for
source and target sentences. Differences in vocabulary,
pragmatics and other linguistic properties all influence the generation of the latent
representations. Therefore, we consider the latent
representations from a different perspective as follows. We
can view
$\boldsymbol{z}_x$ and $\boldsymbol{z}_y$ as
expressions of the sentence-level representations in two
distinct languages based on the same semantics
$\boldsymbol{\epsilon}$ where $\boldsymbol{\epsilon}$ is truly
language-agnostic. $\boldsymbol{z}_x$ and
$\boldsymbol{z}_y$ are obtained by applying parameter-free
techniques such as pooling to the output of the encoder fed
with source and target languages (see
Section \ref{flowUMNT} for details).

\myparagraph{Modeling by NFs.}  For our unsupervised scenario,
we propose to explicitly model the distributions of the
sentence-level representations of both source and target
sentences -- i.e., $p_{\boldsymbol{z}_x}(\boldsymbol{z}_x)$ and
$p_{\boldsymbol{z}_y}(\boldsymbol{z}_y)$ -- using NFs with $K$ sequential flows:
\begin{equation}
p_{\boldsymbol{z}_x}(\boldsymbol{z}_x) = p_{\boldsymbol{\epsilon}}(\boldsymbol{\epsilon}) \prod_{i=1}^{K} \left| \det \frac{\partial f_x^{(i)}(\boldsymbol{z}^{(i)})}{\partial \boldsymbol{z}^{(i)}} \right|^{-1}
\label{equation_srcflow}
\end{equation}
\begin{equation}
p_{\boldsymbol{z}_y}(\boldsymbol{z}_y) = p_{\boldsymbol{\epsilon}}(\boldsymbol{\epsilon}) \prod_{i=1}^{K} \left| \det \frac{\partial f_y^{(i)}(\boldsymbol{z}^{(i)})}{\partial \boldsymbol{z}^{(i)}} \right|^{-1}
\label{equation_tgtflow}
\end{equation}
where $p_{\boldsymbol{\epsilon}}(\boldsymbol{\epsilon})$ is
a base distribution, e.g., the standard normal distribution;
$f_x^{(i)}$ and $f_y^{(i)}$ are the $i^{th}$ transformations
for the source and target languages, respectively; and
$\boldsymbol{z}^{(i)}$ is the intermediate variable where we
define
$\boldsymbol{z}^{(1)} = \boldsymbol{\epsilon}$ and
$\boldsymbol{z}^{(K)} = \boldsymbol{z}_x$ or
$\boldsymbol{z}_y$ for notational convenience. The base distribution can be viewed as
the ``true'' underlying semantic space, abstracting away from
language specifics.

\begin{figure*}
  \centering
  \includegraphics[width=1.0\textwidth]{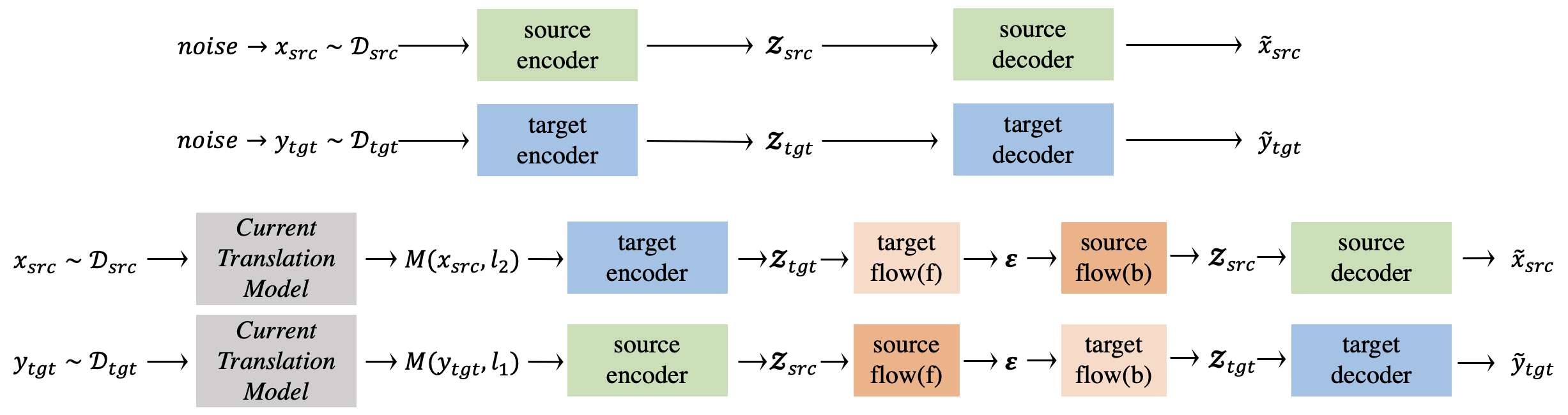}
  \caption{Top two diagrams: denoising auto-encoding for source and
  target sentences.
  Bottom two diagrams: iterative back-translation for
  source and target sentences.
  $M(\boldsymbol{x}, l_2)$
(resp.\  $M(\boldsymbol{y}, l_1)$)
  denotes the
  target-language
  (resp.\ source-language)
  sentence generated by applying the
  current translation model $M$ to
  the source-language  sentence $\boldsymbol{x}$
  (resp.\ the target-language sentence $\boldsymbol{y}$). $l_1$ (the source language) and $l_2$ (the target language) are the parameters specifying the aim of the translation direction of $M$. $\boldsymbol{\Tilde{x}}_{src}$ (resp. $\boldsymbol{\Tilde{y}}_{tgt}$) is the reconstruction of $\boldsymbol{x}_{src}$ (resp. $\boldsymbol{y}_{tgt}$). (f) indicates the flow transforms forward from $\boldsymbol{z}$ to $\boldsymbol{\epsilon}$ while (b) for backward transformation, i.e., from $\boldsymbol{\epsilon}$ to $\boldsymbol{z}$.
  }
  \label{fig:dea_bt}
\end{figure*}

Our transformation to the sentence-level representations is
similar to \cite{li2020sentenceBERT}. They argued that BERT
induces a non-smooth anisotropic semantic space of
sentences, which can harm its accurate representation of semantic
similarity. Therefore, they also used NFs to transform the
anisotropic BERT sentence-level distribution to a standard
Gaussian distribution that is smooth and isotropic and
reported better performance on some sentence-level
similarity tasks. By using this type of sentence-level representation, the semantics of sentences from different languages can therefore be aligned in a simple common space in an unsupervised way, which we show is effective for unsupervised MT.
\enote{hs}{since there is virtually no difference between
the two appraoches, hyou may want to remove th elast
sentence and instead say somethig like:
We show that this type of sentence representation is
effective in unsupervised MT.}
\enote{yl}{sure, I have updated it.}

For
simplicity, we denote the NFs for transforming the distributions of source and target
sentence-level representations to the base distribution as mappings
$G_{(\boldsymbol{z}_x \rightarrow \boldsymbol{\epsilon})}$
and
$G_{(\boldsymbol{z}_y \rightarrow \boldsymbol{\epsilon})}$. Because
of the invertibility property of NFs, these mappings are also
invertible, and we  have
$ G_{(\boldsymbol{\epsilon} \rightarrow \boldsymbol{z}_x)} =
G_{(\boldsymbol{z}_x \rightarrow \boldsymbol{\epsilon})}^{-1}$ and
$ G_{(\boldsymbol{\epsilon} \rightarrow \boldsymbol{z}_y)} =
G_{(\boldsymbol{z}_y \rightarrow \boldsymbol{\epsilon})}^{-1}$.


\myparagraph{Latent Code Transformation.} Inspired by
AlignFlow \citep{grover2020alignflow}, we consider the
cross-domain transformation between $\boldsymbol{z}_x $ and
$\boldsymbol{z}_y$. In this way, we can formulate a
language-specific latent code for the decoder.
We formalize the
cross-language latent code transformation from the source to the target language  as follows:
\begin{equation}
G_{(\boldsymbol{z}_x \rightarrow \boldsymbol{z}_y)} = G_{(\boldsymbol{\epsilon} \rightarrow \boldsymbol{z}_y)} \circ G_{(\boldsymbol{z}_x \rightarrow \boldsymbol{\epsilon})}
\end{equation}
The target-to-source latent code transformation is then the composition of $G_{(\boldsymbol{\epsilon} \rightarrow \boldsymbol{z}_x)}$ and $G_{(\boldsymbol{z}_y \rightarrow \boldsymbol{\epsilon})}$. As $G_{(\boldsymbol{\epsilon} \rightarrow \boldsymbol{z}_y)}$ and $G_{(\boldsymbol{\epsilon} \rightarrow \boldsymbol{z}_x)}$ are the inverse mappings of $G_{(\boldsymbol{z}_y \rightarrow \boldsymbol{\epsilon})}$ and $ G_{(\boldsymbol{z}_y \rightarrow \boldsymbol{\epsilon})}$, we can easily obtain them with normalizing flows, such as realNVP  \citep{dinh2017realnvp} and Glow \citep{Kingma2018Glow}. We also note that $G_{(\boldsymbol{z}_x \rightarrow \boldsymbol{z}_y)}$ and $G_{(\boldsymbol{z}_y \rightarrow \boldsymbol{z}_x)}$ are both invertible since they are compositions of two invertible mappings. Moreover, $G_{(\boldsymbol{z}_x \rightarrow \boldsymbol{z}_y)}$ is the inverse of $G_{(\boldsymbol{z}_y \rightarrow \boldsymbol{z}_x)}$ and vice versa (see Appendix \ref{invertibility} for details).

\subsection{Flow-Adapter Based Unsupervised Machine Translation}\label{flowUMNT}
The general architecture is shown in
Figure \ref{fig:flow-adapter_structure}. The transformer
architecture \citep{vaswani2017attention} is used for both
encoder and decoder. We use source encoder/decoder to denote
the encoder/decoder for encoding/generating the
source-language sentence. Similarly, target encoder/decoder
refer to the encoder/decoder encoding/generating the target-language sentence. The decoders work in an autoregressive way. Source flow and target flow are NFs for modeling the sentence-level latent representations of the source and target language, respectively, as introduced in Section \ref{latentcode}.

\myparagraph{Encoding.} The source encoder and the target
encoder work in the same way; for brevity, we only describe
the procedure of encoding the source sentence and how
$\boldsymbol{z}_x$ is generated. The source encoder takes
the source sentence $\boldsymbol{x}=\{x_0, \cdots, x_S\}$ as
input and generates the hidden representations
$\{\boldsymbol{h}_0, \cdots, \boldsymbol{h}_S\}$. These
hidden representations will be used as encoder-decoder
attentional inputs. In addition, we use the hidden
representations to generate a sentence-level representation
for the source sentence by applying
max-pooling and mean-pooling to the token-level representations. After that, we sum up the results with the CLS representation $\boldsymbol{h}_0$, which usually encodes some global information. Finally, we use a projection matrix $\boldsymbol{W}$ to project the resulting vector to a latent space. The output is referred to as $\boldsymbol{z}_x$, i.e., the sentence-level representation of the source sentence (see Appendix \ref{gslr} for equation and illustration).

\myparagraph{Cross-lingual Translation.}
We hypothesize that the decoder can better leverage
language-specific latent representations (i.e.,
$\boldsymbol{z}_x$ for the source decoder and
$\boldsymbol{z}_y$ for the target decoder) than
indiscriminately using
the same representational space for source and target,
e.g., $\boldsymbol{z}_x$ for the target decoder. Therefore, we
propose to perform a latent code transformation for
cross-language translation as shown in
Figure \ref{fig:flow-adapter_structure}. If the
model is performing the translation in the
source-to-target direction, the source flow first transforms
the source latent representation $\boldsymbol{z}_x$ into
$\boldsymbol{\epsilon}$, which is a vector in the semantic
base space. Then the target flow transforms
$\boldsymbol{\epsilon}$ back into $\boldsymbol{z}_y$, which
is in the target latent representation space. Then
$\boldsymbol{z}_y$ is used in the target decoder for generating the target sentence.

\myparagraph{Denoising Auto-Encoding (DAE) and Back Translation (BT)
Processes.} The DAE reconstructs a sentence from its noised version. For inducing noise, we use the same strategy which is used by \cite{LampleCDR18umtmco} (For more details, please refer to Appendix \ref{hyperparam}). Since we train the DAEs separately for source and target languages, hence we don't need a latent code transformation there. For BT, however, such a latent code transformation is performed twice; taking BT for the source language as an example: first in the source-to-target direction, then in the target-to-source direction as shown in Figure \ref{fig:dea_bt}.
\enote{hj}{Can we can a line about the noise augmentation we used? Was it the same as XLM?}
\enote{yl}{the noise we used is the same as XLM, the detail about the noise is omitted in the main context. I put the detail into the appendix.}

\myparagraph{Decoding.}  To enable the decoder to capture the global semantics and mitigate improper alignments, we use the procedure outlined in \cite{Hendra2020variational}, and incorporate the latent representation $\boldsymbol{z}$ into the output of the last layer of the decoder $\{\boldsymbol{s}_0, \cdots, \boldsymbol{s}_T\}$:
\begin{equation}
\boldsymbol{o}_i = (1 - \boldsymbol{g}_i) \odot \boldsymbol{s}_i + \boldsymbol{g}_i \odot  \boldsymbol{z}
\end{equation}
where $\boldsymbol{g}_i
= \sigma([\boldsymbol{s}_i; \boldsymbol{z}])$, $\sigma
(\cdot)$ is the sigmoid function, $\odot$ denotes Hadamard
product between two vectors, and $\boldsymbol{o}_i$ is the
logit vector used to generate a prediction at the $i^{th}$
position. The values in $\boldsymbol{g}_i$ control the
contribution of $\boldsymbol{z}$ to $\boldsymbol{o}_i$. In
case the dimension of the latent representation does not
match the dimension of the decoder output, a linear
projection maps $\boldsymbol{z}$ to the desired dimension.

\myparagraph{Training.}
Our flow-adapter framework has three learning objectives:
DAE, BT and MLE of the sentence-level
representations. The description of DAE and BT is omitted
here as they are well known from  related
work \citep{LampleCDR18umtmco, ArtetxeLAC18unmt}. A single training iteration consists of a DAE step followed by a BT step as shown in Figure \ref{fig:dea_bt}. MLE computation is integrated into the DAE step to calculate the likelihood of the sentence-level representations. Our MLE learning objective for the source monolingual dataset can be
formulated as follows (similar for the target dataset, omitted):
\begin{equation}
\mathcal{L}_{MLE}(G_{(\boldsymbol{z}_x \rightarrow \boldsymbol{\epsilon})}) = E_{z \sim p_{\boldsymbol{z}_x}} [\log p_{\boldsymbol{z}_x}(z)]
\end{equation}
where
\begin{equation}
p_{\boldsymbol{z}_x}(z) = p_{\boldsymbol {\epsilon}}( G_{(\boldsymbol{z}_x \rightarrow \boldsymbol{\epsilon})}(z)) \left| \det \frac{\partial G_{(\boldsymbol{z}_x \rightarrow \boldsymbol{\epsilon})}}{\partial \boldsymbol{z}_x} \right|_{\boldsymbol{z}_x =z}
\end{equation}
by definition of the source NFs in Equation \ref{equation_srcflow}. $E_{z \sim p_{\boldsymbol{z}_x}}$ is approximated via mini-batches of sentence-level latent representations generated by the encoder in the training process. By training the source flow and the target flow with this MLE loss, the flows can therefore transform between the language-specific latent space of the representations and the base semantic space. In this way, the latent code transformations, i.e., $G_{(\boldsymbol{z}_x \rightarrow \boldsymbol{z}_y)}$ and $G_{(\boldsymbol{z}_y \rightarrow \boldsymbol{z}_x)}$ can be constructed.

\section{Experiments}
\subsection{Datasets}
\myparagraph{Multi30K task1 dataset
\citep{elliott2016multi30k,elliott2017findings}.\footnote{https://github.com/multi30k/dataset}}
This is a multi-modal dataset that has 30,000 images
annotated with captions in English, German and French. Similar to \citep{LampleCDR18umtmco}, we only
use the caption of each image. The officially provided
train, validation and test sets are used. We use this dataset
as a small-scale test for validating the
effectiveness of our methods.

\myparagraph{WMT datasets.\footnote{http://www.statmt.org/}}
Our experiments are run with the settings that were used for
XLM \citep{Conneau2019xlm}. XLM uses the monolingual data
from the WMT News Crawl datasets\footnote{https://github.com/facebookresearch/XLM/blob/main/get-data-nmt.sh}. We report results on \textit{newstest2014 en-fr}, \textit{newstest2016 en-de} and \textit{newstest2016 en-ro}.

\enote{hs}{edited above paragraph, check}
\enote{yl}{yes, this is ok.}


\subsection{Setups}
\myparagraph{Preprocessing.} We tokenize the sentences with
the Moses script \citep{koehn2007moses}. For the Multi30K
dataset, we process it similar
to \citet{LampleCDR18umtmco}. Specifically, the sentences
are randomly divided into two parts. The source-language
monolingual dataset is built from the source-language
sentences in the first part and the target-language dataset from
the second part. In this way, there will be no exact
translations of any sentences in the datasets. For the WMT datasets,
we use the preprocessing methods
from \citep{Conneau2019xlm}. For the English-Romanian dataset,
we remove the diacritics as done
by \citet{sennrich2016edinburgh} to avoid their inconsistent
usage in the Romanian part of the dataset.

\myparagraph{Metric \& Performance.} We use BLEU as
metric \citep{papineni2002bleu} for all our
experiments. Although \citet{artetxe2020rigorunsuo}
recommended to use unsupervised validation criteria for
systematic tuning, we follow the setting
of \citep{Conneau2019xlm,SongTQLL19mass} and use the provided parallel validation sets for tuning hyperparameters. We report the results on the test sets of the models that achieve
best performance on the validation sets.

\myparagraph{Pre-trained Embeddings \& Models.}
We use the pre-trained
MUSE\footnote{https://github.com/facebookresearch/MUSE} \citep{LampleCRDJ18wordtrans}
embeddings for the multilingual unsupervised MT experiment (Table \ref{tab:multi_trans_perf}). We also leverage 
pre-trained cross-lingual models in the experiment of shared \& separate decoder(s) (Table \ref{tab:flow_pretrained_perf}). Specifically, XLM models from
HuggingFace\footnote{https://github.com/huggingface} \citep{wolf2020transformers}
are used to initialize the encoder. Moreover, we also incorporate our 
flow-adapter architecture directly into the codebase of the original
implementation of
XLM\footnote{https://github.com/facebookresearch/XLM} for the WMT dataset experiment (Table \ref{tab:wmt_trans_perf2}).
In this case,
the encoder and decoder are both initialized with pre-trained models. Details of these models can be found in Appendix \ref{hyperparam}.

\def\tmpsep{0.2cm}

\begin{table}
\centering
\footnotesize
\begin{tabular}{l@{\hspace{\tmpsep}}l@{\hspace{\tmpsep}}l@{\hspace{\tmpsep}}l@{\hspace{\tmpsep}}l@{\hspace{\tmpsep}}l@{\hspace{\tmpsep}}l}
\hline
Models & en-de & de-en & en-fr & fr-en & de-fr & fr-de\\
\hline
baseline & 11.87 & 19.31 & 16.52 & 19.24 & 11.03 & 8.36\\
\textbf{3-scf} & \textbf{12.25} & 19.83 & \textbf{16.98} & \textbf{20.12} & \textbf{11.67} & \textbf{8.98}\\
\textbf{3-glow} & 11.91 & \textbf{20.14} & 16.86 & 19.55 & 11.49 & 8.61\\
\hline
\end{tabular}
\caption{\label{tab:multi_trans_perf}
BLEU of our flow-adapter model for
multilingual translation on Multi30K.
Baseline: model without our proposed flow-adapter
architecture. 3-scf or 3-glow models:
baseline models
with the flow-adapter architecture constructed by two realNVP NF models or Glow NF models, each of which consists of 3
sequential flows, for performing the latent code transformation in that translation direction.}
\end{table}

\subsection{Results of Multilingual Unsupervised Machine Translation on Multi30K}\label{sbsection:multilingual}

As Multi30K  provides parallel test data for English,
French and German, we first conduct
experiments to show the multilingual translation ability of
our flow-adapter models.
The
results are shown in Table \ref{tab:multi_trans_perf}. The
baseline model (without flow-adapter architecture) is trained with only DAE loss, while the
flow-adapter based models (3-scf and 3-glow) are additionally trained with MLE
loss for the NFs. 3-scf (resp.\ 3-glow) is the baseline
model with two realNVP NF models \citep{dinh2017realnvp}
(resp.\ Glow NF models \citep{Kingma2018Glow})
, each of which consists of 3 sequential flows. Each NF model is used to model the sentence-level representations of one specific language, and two NF models then construct a flow-adapter for that translation direction (as shown in Figure \ref{fig:flow-adapter_structure}).
The flow-adapter based
models additionally perform the latent code transformation to
generate a language-specific representation while the baseline model does not perform such a transformation. 

\enote{hs}{above: why ``two'' realNVP flows? where does the
number two come from?}
\enote{yl}{"two" refers to the NF models involved in that translation direction. For example, en-de translation will use the NF model of English and NF model of German to construct the flow-adapter. I have updated the above sentences.}

For this experiment, we use the pre-trained
cross-lingual word embeddings (MUSE embeddings) and randomly initialize a
shared encoder and a shared decoder for all three
languages. It is worth noting that the training objective
does not contain the iterative back-translation. For further
research where there are far more languages accommodated,
random online back-translation (ROBT) proposed
by \citet{zhang2020improving} could be considered.

Table \ref{tab:multi_trans_perf} shows
improvements over all six translation
directions by using the flow-adapter architecture. Notably,
our 3-scf and 3-glow models achieve 19.83 and 20.14 BLEU
scores, respectively, on \textit{de-en}, which is 0.52
and 0.83 higher than the baseline model. Similar
improvements can also be seen on other translation
directions. Our 3-scf model
has BLEU scores that are 0.46 to 0.88 higher than the baselines
while our 3-glow model has BLEU scores that are
0.04 to 0.83 higher than the baselines.
The overall improvements show that the flow-adapter
can generate more suitable sentence-level
representations by performing the latent code
transformation, which is helpful for the decoder to
capture the semantics and generate more suitable
translations.

\begin{table*}
\centering
\footnotesize
\begin{tabular}{lrrrr}
\hline
Models & en-de & de-en & en-fr & fr-en \\
\hline
baseline (shared decoder) & 0.25 & 0.17 & 0.13 & 0.11\\
\textbf{3-scf} (shared decoder) & 25.80 & 28.92 & 39.26 & \textbf{36.84}\\
\textbf{3-glow} (shared decoder) & 26.09 & 29.48 & 39.21 & 36.66\\
baseline (separate decoders) & 27.54 & 28.97 & 39.17 & 36.27\\
\textbf{3-scf} (separate decoders) & 28.24 & \textbf{30.63} & \textbf{39.64} & 36.45\\
\textbf{3-glow} (separate decoders) & \textbf{28.79} & 30.45 & 39.31 & 36.29\\
\hline
UNMT \citep{LampleCDR18umtmco} & 22.74 & 26.26 & 32.76 & 32.07 \\
\hline
\end{tabular}
\caption{\label{tab:flow_pretrained_perf}
BLEU of the flow-adapter  models and 
unsupervised SOTA model, i.e., UNMT \citep{LampleCDR18umtmco},
on Multi30K.
Baseline models use pre-trained XLMs from HuggingFace as the encoder and randomly initialized decoder(s) without the flow-adapter. (separate decoders): two independent and
randomly initialized decoders are used, each for decoding
a specific language. (shared decoder): a single
shared decoder for decoding both
languages is used. 3-scf and 3-glow (as defined in Table \ref{tab:multi_trans_perf} and Section \ref{sbsection:multilingual}) denote the baseline models with the flow-adapter architecture. We report the results of UNMT from their original
paper.}
\end{table*}

\enote{hs}{caption: when you mention state of the art,
ideally, you should give a reference: which paepr are you
referring to?}
\enote{yl}{the SOTA refers to UNMT paper. I have updated it.}

We also find that the translation performance
is closely related to the language pair and the translation
direction for both the baseline models and flow-adapter models. Our models obtain very good performance
on \textit{en-fr}, with performances in
both the \textit{en-fr} or \textit{fr-en} directions better
by 16 BLEU points. For other language pairs
(including \textit{en-fr}), there is always one direction
showing better performance than the other.
Specifically, \textit{de-en} achieves more
than 19 BLEU points compared with 12 points for \textit{en-de},
and \textit{de-fr} achieves more than 11 BLEU
points compared with 8.5 for \textit{fr-de}.
\enote{hs}{the following sounds weird: it can be said about
a vast majority of MT papers. it sounds like someone who
just entered the MT field is stating something that has been
known for decades}
\enote{yl}{so maybe we should delete the following statements?}

\subsection{Results of Shared-Decoder \& Separate-Decoder Models on Multi30K}\label{sbsection:decoders}
We present the performance of our flow-adapter
models under the shared-decoder and separate-decoder
settings on Multi30K. For this experiment, the encoder is initialized
with the pre-trained XLM model and fixed; the decoder
parameters are 
randomly initialized and then trained. We also report
the performance of a previous SOTA model, i.e., UNMT \citep{LampleCDR18umtmco}.\footnote{UNMT did not use pre-trained models. The results are therefore not strictly comparable to ours.} The results are
shown in Table \ref{tab:flow_pretrained_perf}. First, we
notice that the shared-decoder baseline model obtains very
low BLEU scores. By checking the translation generated, we
find the model only copies the input as translation. This
phenomenon shows that this baseline, which
does not perform the latent code transformation, cannot
model two languages simultaneously well, and thus cannot
generate translations in the desired language domains. However,
by incorporating the flow-adapter, the models will no longer
have this limitation. Both shared-decoder 
models, i.e., 3-scf and 3-glow, achieve very good performance on all translation
pairs. For example, the 3-scf model obtains
BLEU scores of
25.80, 28.92,
39.26 and 36.84
on \textit{en-de}, \textit{de-en}, \textit{en-fr}
and \textit{fr-en}, which are 
much higher than the baseline.

Compared with the shared-decoder scenario, the models under
the separate-decoder setting do not suffer from the copying problem,
because different decoders are used to specifically model
and generate sentences in distinct language domains. The
downside, however, is that using multiple decoders at the
same time can substantially increase the number of trainable
parameters. Within the separate-decoder models, the
flow-adapter models generally perform better than the
baseline model, with about 1 BLEU increase on \textit{en-de}
and \textit{de-en} directions and relatively smaller improvements
on \textit{en-fr} and \textit{fr-en}. Those improvements
demonstrate that the model can benefit from the flow-adapter
architectures as language-specific latent representations
are used, thus advancing the translation quality.

We also observe that the separate-decoder models generally
perform better than the shared-decoder models. The
separate-decoder baseline is much better than its
counterpart as it avoids the copying problem. For the 3-scf
flow-adapter models, we find that the separate-decoder
model outperforms the shared-decoder model by 2.44, 1.71, 0.38
on \textit{en-de}, \textit{de-en} and \textit{en-fr}.
However, on \textit{fr-en}, the shared-decoder
model achieves a BLEU socre that is by 0.39 BLEU points
better. A similar phenomenon
can also be seen for the 3-glow model. We conjecture this
is due to the
similarity between languages. As English and French share
common vocabulary, some common features can therefore be
captured by a shared decoder, thus improving its
generalization.

Lastly, when compared with UNMT, our models show
superiority, improving performance by more than 4 BLEU
points in each direction. We attribute the improvements to
the usage of the pre-trained model and incorporation of
language-specific sentence-level representations obtained by
our latent code transformation.

\begin{table*}
\centering
\footnotesize
\begin{tabular}{lrrrrrr}
\hline
Models & en-de & de-en & en-ro & ro-en & en-fr & fr-en\\
\hline
XLM (EMD + EMD) \citep{Conneau2019xlm}& 21.30 & 27.30 & 27.50 & 26.60 & 29.40 & 29.40\\
XLM (MLM + MLM) \citep{Conneau2019xlm}& 26.40 & \textbf{34.30} & 33.30 & \textbf{31.80} & 33.40 & 33.30\\
\textbf{5-scf} & \textbf{26.50} & 32.63 & \textbf{34.11} & 31.69 & \textbf{35.77} & \textbf{33.72}\\
\textbf{5-glow} & 26.43 & 32.04 & 33.87 & 31.32 & 35.25 & 33.12\\
\hline
MASS \citep{SongTQLL19mass}& 28.30 & 35.20 & 35.20 & 33.10 & 37.50 & 34.90 \\
CSP and fine-tuning \citep{yang2020code}& 28.70 & 35.70 & - & - & 37.90 & 34.50 \\
\hline
\end{tabular}
\caption{\label{tab:wmt_trans_perf2}
BLEU of the flow-adapter models (5-scf and 5-glow) and SOTA
models on WMT datasets. XLM (MLM + MLM) is the baseline in
this table, as 5-scf and 5-glow
use it as the base model for
the flow-adapter architecture. We report the results of
XLM, MASS and CSP from the original paper.}
\end{table*}

\subsection{Results on WMT datasets}

We further integrate our flow-adapter architecture into the
original implementation of XLM \citep{Conneau2019xlm} and
conduct experiments on the WMT datasets. To fully leverage the
pre-trained models, we initialize both the encoder and
decoder with XLM models and set them trainable. In contrast to
the  experiment in
Section \ref{sbsection:decoders}, a single shared decoder is
used for this experiment, since the decoder is also
initialized with the pre-trained model and has far more
parameters compared with the randomly initialized
transformer decoder we use in Section \ref{sbsection:decoders}. We
report the performance of the flow-adapter based models (5-scf and
5-glow\footnote{Preliminary experiments showed that using 5 flows
provides slightly better results than 3 flows for WMT as WMT has many more sentences than
Multi30K and therefore more powerful generative models (by adding more intermediate flows) are
needed to model the sentence-level representations.}) as well as the performance of the SOTA models,
namely XLM, MASS and CSP.\footnote{We follow  prior
convention
and compare directly with MASS and CSP even though dataset
processing for MASS and CSP (e.g., filtering, sampling) are
not strictly the same as for XLM. But the difference is small and results would not be much different as \citet{yang2020code} mentions.} The results are shown in
Table \ref{tab:wmt_trans_perf2}. Noticeably, both of our
flow-adapter  models achieve remarkable
performance on all language pairs. Compared with the results
of XLM (EMD + EMD), which uses the pre-trained cross-lingual
embeddings instead of pre-trained models, both  5-scf and
5-glow  achieve overall better performance. For
example, 3-scf achieves BLEU scores higher by 5.20, 5.33, 6.61, 5.09, 6.37 and
4.32 
on \textit{en-de}, \textit{de-en}, \textit{en-ro}, \textit{ro-en}, \textit{en-fr}
and \textit{fr-en}, respectively. While not being as
good as 5-scf, 5-glow still shows superiority over XLM (EMD +
EMD). These improvements can be contributed to (1) the usage
of pre-trained models and (2) the introduction of the
flow-adapter.

We further compare our flow-adapter based models with XLM
(MLM + MLM), which is also initialized with pre-trained
models. We find the performance of \textit{x-en} directions
is consistently lower than \textit{en-x} directions for our
models except for \textit{en-de}. This pattern is not
limited to our architecture but is consistently present in
prior work. We, again,
speculate this is relating to the complexity of languages as
well as similarity between languages. We leave this finding
for future investigation. Our flow-adapter based models,
though achieving similar or relatively worse BLEU scores
on \textit{de-en} and \textit{ro-en} compared with XLM (MLM
+ MLM), obtain higher scores on other directions,
i.e., \textit{en-de} and \textit{en-ro}, suggesting that our
models  might be more helpful on specific translation
directions, as the flow-adapter generates language-specific
representations. Lastly,  5-scf achieves scores by 2.37 and 0.42
 better than XLM (MLM + MLM) on \textit{en-fr}
and \textit{fr-en}. As in the other experiments,
the improvement due to flow adapters seems to be related to the
 languages involved in that language pair and the translation
directions. We would like to investigate this phenomenon in
future research.

Finally, out models are competitve  with MASS and CSP, with only small differences in BLEU.
In general,
the experiments
shows the validity and effectiveness of our
 flow-adapter architecture integrated into
pre-trained models.

\section{Conclusion}
In this work, we propose a novel flow-adapter architecture
for unsupervised NMT. The flow-adapter employs a pair of NFs
to explicitly model the distributions of the sentence-level
representations. A latent code transformation is performed
in translation, which enables the decoder to better capture
the semantics of sentences in certain language
domains. Through extensive experiments, we show the
flow-adapter can improve multilingual translation
ability. Moreover, it can alleviate the copying problem. By
integrating the flow-adapter into pre-trained XLM models, we
achieve results competitive to state-of-the-art models on WMT
datasets.

In the future, we would like to explore the possibility of
pre-training the flow-adapter simultaneously when
pre-training the language models so that the flows can learn
more information. Moreover, we would like to extend
normalizing flows to language generation. By using different
flows for different languages, multilingual language
generation of the same semantics can be performed.

\enote{hs}{I would cut this. this is well-trodden territory
for anybody who knows MT. it sounds like somebody who just
started in MT is speculating without really knowing the vast
body of literature that is releavnt to these questions:
We discovered some interesting patterns in our results that
pertain to the scores in various translation directions (en
$\rightarrow$ de vs.\ de $\rightarrow$ en etc). From
Table \ref{tab:wmt_trans_perf2} we can see that
for \textbf{all} listed architectures, the BLEU scores are
better from de $\rightarrow$ en (compared to en
$\rightarrow$ de), and worse in ro $\rightarrow$ en and fr
$\rightarrow$ en (compared to their respective opposite
directions). We emphasize that since these patterns are not
limited to our architecture but common to other recent UNMT
architectures, this suggests that linguistic phenomenona
rather than properties of the architecture are the cause and
hence a separate investigation into its nature and origin
may be required.}
\enote{hj}{So I just had a meeting with Alexandra and she was also surprised at these
patterns. Though it's not clear if MT community has worked on these patterns or not. Maybe Alex would have some comments.}

\enote{hs}{instead of destabilizing the current draft, I
would cut this -- there is actually an argument to be made
that you shouldn't give away your best idea before you've
had a chance to publish them yourself:
In the future, we would like to explore the possibility of
pre-training the flow-adapter simultaneously when
pre-training the language models so that the flows can learn
more information. Moreover, we would like to extend
normalizing flows to language generation. By using different
flows for different languages, multilingual language
generation of the same semantics can be performed.
}
\enote{yl}{sure, it makes sense :)}
\enote{hj}{Added that part.}
\enote{yl}{if adding this part now the text is more than 9 pages, revision is needed.}

\section*{Acknowledgements}
We are grateful to Alex Fraser and Alexandra Chronopoulou for their insightful input. This work was funded by the European Research Council (ERC \#740516).

\bibliography{anthology,custom}
\bibliographystyle{acl_natbib}

\appendix

\section{Appendix}
\label{sec:appendix}

\subsection{Proof of the Invertibility}\label{invertibility}

The following proof is based on the proof by \citet{grover2020alignflow} and shows the source-to-target latent code transformation is the inverse of the target-to-source latent code transformation, and vice versa:
\begin{equation}
\begin{aligned}
G_{(\boldsymbol{z}_x \rightarrow \boldsymbol{z}_y)}^{-1} &= 
(G_{(\boldsymbol{\epsilon} \rightarrow \boldsymbol{z}_y)} \circ G_{(\boldsymbol{z}_x \rightarrow \boldsymbol{\epsilon})})^{-1}\\ 
&= G_{(\boldsymbol{z}_x \rightarrow \boldsymbol{\epsilon})}^{-1} \circ G_{(\boldsymbol{\epsilon} \rightarrow \boldsymbol{z}_y)}^{-1} \\ 
&= G_{(\boldsymbol{\epsilon} \rightarrow \boldsymbol{z}_x)} \circ G_{(\boldsymbol{z}_y \rightarrow \boldsymbol{\epsilon})} \\ 
&= G_{(\boldsymbol{z}_y \rightarrow \boldsymbol{z}_x)}
\end{aligned}
\end{equation}

\subsection{Generation of Sentence-level Representation}\label{gslr}

The following formula shows the process of how the sentence-level representation is generated:

\begin{figure*}
  \centering
  \includegraphics[width=1.0\textwidth]{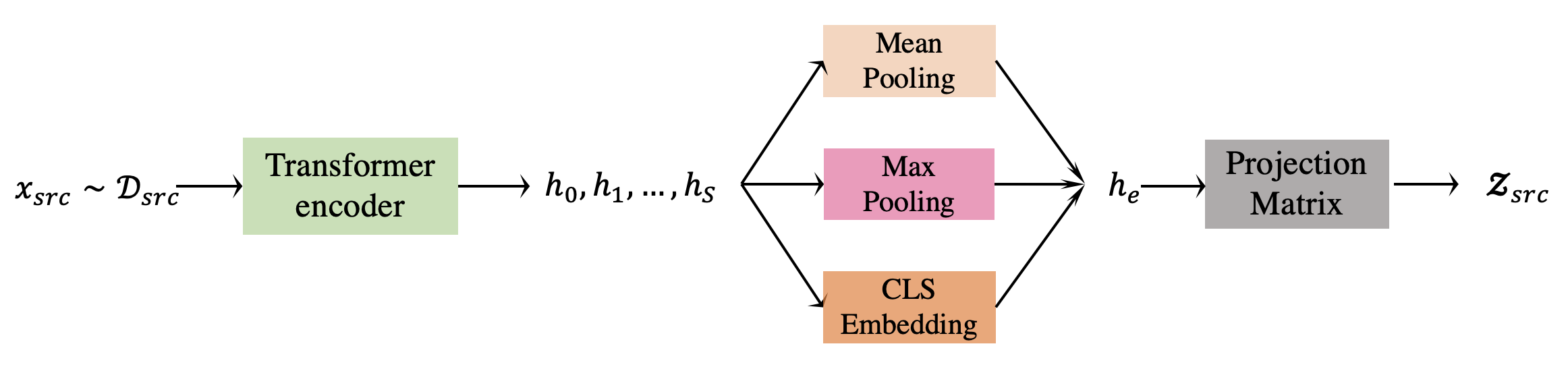}
  \caption{The illustration of generation of the sentence-level representations. CLS embedding refers to the first vector output by the transformer encoder, i.e., $\boldsymbol{h}_0$.}
  \label{fig:gsr}
\end{figure*}

\begin{equation}
\begin{aligned}
\boldsymbol{z} = \text{Linear} &(\text{max-pool}([\boldsymbol{h}_0, \cdots, \boldsymbol{h}_S]) \\
&+ \text{mean-pool}([\boldsymbol{h}_0, \cdots, \boldsymbol{h}_S]) \\
&+ \boldsymbol{h}_0)
\end{aligned}
\end{equation}
where the pooling operation generates a vector that has the same dimension as $\boldsymbol{h}_0$, so the three vectors have the same shape and therefore are additive. An illustration can be seen in Figure \ref{fig:gsr}.

\subsection{Details of the Experiments: Models \& Hyperparameters}\label{hyperparam}

\subsubsection{Multi30K Experiment}
For the multilingual machine translation tasks, we use the cross-lingual MUSE \citep{LampleCRDJ18wordtrans}. The embeddings were
learned using
fastText\footnote{https://github.com/facebookresearch/fastText} \citep{bojanowski2017enriching}
on Wikipedia data and then aligned in a common space by the
method proposed by \citet{LampleCRDJ18wordtrans}. The results shown in Table \ref{tab:multi_trans_perf} is the average over 10 runs on the test sets.
Denosing auto-encoding is used to train the baseline model. The flow-adapter based (3-scf and 3-glow) models are additionally trained with MLE loss. We follow the denoising auto-encoding hyperparameter settings used by \citet{LampleCDR18umtmco}. Specifically, word drop and word shuffling are used. For word drop, every word in a sentence (except <bos> and <eos>) can be dropped with a probability $p_{wd}$, which we set 0.1 in our experiments. For word shuffling, a random permutation $\sigma$ is applied to the input sentence, which satisfy the condition: $\forall i \in \{1, n\}, |\sigma(i) - i| \le k$, where $i$ is the index of a word in the sequence, $n$ is the length of the sequence and $k$ is a hyperparameter that controls the degree of the permutation which we set 3 in our experiments. The dimension of the pre-trained embedding is 300. The randomly initialized shared encoder and decoder use transformer architecture with 512 hidden units, 4 heads and 3 layers by default. We use separate embedding layers for each language and tie their weights with the output layers for each language. The size of the sentence-level latent representation is set to 100. And the weight of the MLE loss for the flows is set to 0.01. We use dropout \citep{srivastava2014dropout} probability of 0.2 for the transformers and 0 for the flows. The batch size is set to 32. The whole model is trained in an end-to-end manner with Adam optimizer \citep{kingma2015adam} with an initial learning rate of 0.0001.

For the shared-decoder \& separate-decoder experiments, we
use the pre-trained language
models \textit{xlm-mlm-enfr-1024}, \textit{xlm-mlm-ende-1024}, \textit{xlm-mlm-enro-1024}
from
HuggingFace\footnote{https://github.com/huggingface} \citep{wolf2020transformers}
to initialize a shared encoder and randomly initialize the
decoder(s).using pre-trained models. Denosing auto-encoding
and iterative back-translation are used to train the
baseline model. The flow-adapter based (3-scf and 3-glow)
models are additionally trained with MLE loss. The same
denoising auto-encoding hyperparameters as above are
used. For iterative back-translation, greedy decoding is
used to generate synthetic parallel sentences as well as the
reconstructions. A single embedding layer (from the
pre-trained model) is used for both the source and target
languages and its weight is tied with the output layer. The
parameters of the encoder are fixed except for its embedding
layer which is also used by the decoder(s). The size of the
sentence-level latent representation is set to 256. The
pre-trained encoder uses 1024 as the embedding size and GELU
activations \citep{dan2016gelu}, and has 4096 hidden units,
8 heads and 6 layers. The randomly initialized decoder has
512 hidden units, 8 heads and 3 layers. The models are
firstly trained with DAE loss (and MLE loss for flow-adapter
models) for the first 3 epochs, then trained with all losses
(including the iterative back-translation) for the rest
epochs. The rest hyperparameters are the same as above.

\subsubsection{WMT Experiment}
We insert our implementation of flow-adapter architecture into the codebase of XLM\footnote{https://github.com/facebookresearch/XLM\#iii-applications-supervised--unsupervised-mt} and use the pre-trained model of \textit{en-fr}, \textit{en-de} and \textit{en-ro} from them. We also follow their recommended unsupervised training settings. For the flow-related hyperparameters, we use 256 as the size of the sentence-level latent representation. The weight of the MLE loss is set to 0.01.

\end{document}